\documentclass{Interspeech}
\usepackage{amsmath,graphicx}
\usepackage{multirow}
\usepackage{xcolor}
\usepackage[final]{changes}

\renewcommand{\replaced}[2]{#2}

\interspeechcameraready

\title{Word Level Timestamp Generation for Automatic Speech Recognition and Translation}

\author[affiliation={1}]{Ke}{Hu}
\author[affiliation={1}]{Krishna}{Puvvada}
\author[affiliation={1}]{Elena}{Rastorgueva}
\author[affiliation={1}]{Zhehuai}{Chen}
\author[affiliation={1}]{He}{Huang}
\author[affiliation={1}]{Shuoyang}{Ding}
\author[affiliation={1}]{Kunal}{Dhawan}
\author[affiliation={1}]{Hainan}{Xu}
\author[affiliation={1}]{Jagadeesh}{Balam}
\author[affiliation={1}]{Boris}{Ginsburg}

\affiliation{}{NVIDIA}{USA}
\email{kevinhu@nvidia.com}

\usepackage{comment}

\begin{document}

\maketitle

\begin{abstract}
    We introduce a data-driven approach for enabling word-level timestamp prediction in the Canary model. Accurate timestamp information is crucial for a variety of downstream tasks such as speech content retrieval and timed subtitles. While traditional hybrid systems and end-to-end (E2E) models may employ external modules for timestamp prediction, our approach eliminates the need for separate alignment mechanisms. By leveraging the NeMo Forced Aligner (NFA) as a teacher model, we generate word-level timestamps and train \replaced{a Multitask}{the Canary} model to predict timestamps directly. We introduce a new \texttt{<|timestamp|>} token, enabling the Canary model to predict start and end timestamps for each word. Our method demonstrates precision and recall rates between 80\% and 90\%, with timestamp prediction errors ranging from 20 to 120 ms across four languages, with minimal WER degradation. Additionally, we extend our system to automatic speech translation (AST) tasks, achieving timestamp prediction errors around 200 milliseconds. Our code is open-sourced through NeMo\footnote{\scriptsize\url{https://github.com/NVIDIA/NeMo}}, and the checkpoint is released at Hugging Face\footnote{\scriptsize\url{https://huggingface.co/nvidia/canary-1b-flash}}.
\end{abstract}

\section{Introduction}

Accurate timing information is highly desirable for automatic speech recognition (ASR) and downstream tasks such as speech content retrieval (e.g., for lookup of speech content by keywords) and generating timed subtitles. In addition, recent works show that transcription with word-level timestamps also facilitates other speech tasks such as speech-to-speech conversation \cite{defossez2024moshi} or decoding with time flow information \cite{seide2024speech}.

There have been a number of works developing timestamp predictions for various ASR models. Traditional hybrid systems \cite{mcauliffe2017montreal}, which are trained with phoneme alignments, naturally provide alignments between speech frames and text. In the E2E framework, although models like  connectionist temporal classification (CTC) are not directly trained with phoneme or word alignments, one can still use Viterbi decoding to the log-probabilities output by CTC models and reference text for forced alignment (e.g., \cite{rastorgueva2023nemo}). Other E2E models perform timestamp prediction as a separate pass in addition to regular decoding. Systems like WhisperX \cite{bain2023whisperx} run voice activity detection and a separate forced phoneme alignment to generate word-level timestamps. WhisperTimestamped and related methods \cite{lintoai2023whispertimestamped, wagner2024crisperwhisper} first generate cross attention weights between speech and predicted words during decoding and then use dynamic time warping to generate a word-level alignment. These methods usually need a long chunk of speech to be available for aligning.

Instead of using a separate module or method to generate timestamps, there have been works on enabling E2E models to directly emit timestamps. For example, forced alignment generated by a conventional ASR model is used to teach RNN Transducer (RNN-T) to emit timestamps \cite{sainath2020emitting} or guide the RNN-T training to emit word tokens at the specified time \cite{zhao2021addressing}. Recently, \cite{xu2023efficient} proposes a time-token transducer to jointly model label and token duration prediction. While the model is efficient in decoding, it may delay the duration emission which leads to inaccurate word-level start and end timestamps. Whisper \cite{radford2023robust} and OWSM \cite{peng2023reproducing} use a data-driven approach to directly train the model to emit timestamps. However, the timestamps are at the \textit{segment level} due to training data limitation. Recently, OpenAI released the feature to generate \textit{word-level} timestamps for Whisper \cite{whisper_word_timstamp}, but it is unclear how the feature is developed. 

\begin{figure*}[t]
    \centering
    \includegraphics[width=0.8\textwidth]{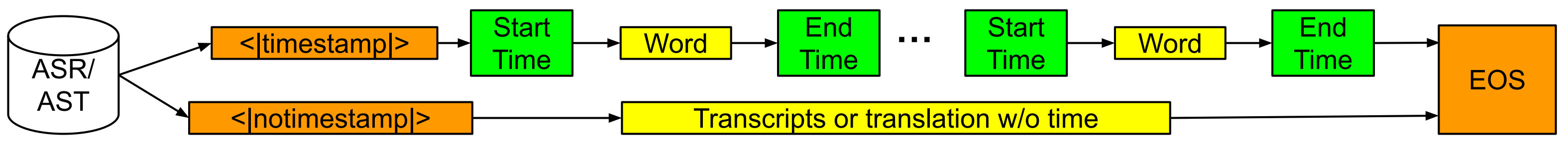}
    \caption{Timestamp training data format. Our training data consists of both ASR or AST training data. Either \texttt{<\!|timestamp\!|>} and \texttt{<\!|notimestamp\!|>} prompt tokens are used for prompting. Both start and end timestamps are added for each word in training.}
    \label{fig:ts}
    \vspace{-1em}
\end{figure*}

In this work, we research a data driven approach to enable word-level timestamp prediction for the Canary model \cite{puvvada2024less}. To overcome the training data scarcity issue, we use the NeMo Forced Aligner (NFA) \cite{rastorgueva2023nemo} as the teacher to generate word-level timestamps for training our timestamp model. We note that the proposed method can work with any teacher data at either word or segment level. To add the timestamp prediction ability, we introduce a new \texttt{<|timestamp|>} prompt for the original Canary model as a new task. Similar to \cite{radford2023robust, peng2023reproducing}, we model timestamps as special tokens up to a maximum duration (i.e., 36 sec in this work). We show that our approach can achieve decent precision and recall (ranging from 80\% to 90\%) for timestamp prediction and the prediction errors ranges from the 20 to 120 ms range for 4 languages, with slight degradation on ASR quality. In addition to ASR, we further extend the system to perform word-level timestamp prediction for automatic speech translation (AST). We define this task as, for each translated word, predicting the start and end frame positions of that word in the original speech. We also follow a data driven approach for this new task. To generate teacher time alignments, we first use NFA to align source speech and source text at the word level, and then use awesome-align \cite{dou2021word} to obtain alignment between source and target text at the word level. The results show that our model can predict word-level timestamps with average errors in the range of 200 ms for 6 language pairs, but this comes with a translation performance drop of 4.4 BLEU points or 2.6 COMET score.

The novelty of our work is two-fold: 1) We propose a solution for multilingual word-level timestamp prediction using teacher generated data. The ASR timestamp prediction has a good coverage of precision and recall at around 90\%, and prediction errors in the range of 50-60 ms on average, and 2) we show that by a data-driven teacher-student learning, the model is also capable of predicting word-level timestamps for speech translation. As far as we know, our work represents the first investigation for word-level timestamp prediction for AST.

\section{Modeling Details}
\label{sec:model}

We use the Canary model \cite{puvvada2024less} as the baseline for adding timestamp ability. \replaced{\cite{puvvada2024less}}{Canary} is a multilingual model which is capable of multiple tasks such as ASR and AST for \replaced{several languages}{English, French, Spanish, and German}. \replaced{\cite{puvvada2024less}}{Canary} models multitasking by using various prompts tokens, such as \texttt{<\!|transcribe\!|>}, \texttt{<\!|translate\!|>}, \texttt{<\!|pnc\!|>}, \texttt{<\!|nopnc\!|>}. To add timestamp prediction ability to \replaced{the Multitask model}{Canary}, we introduce \texttt{<\!|timestamp\!|>} and \texttt{<\!|notimestamp\!|>} for generating word-level timestamps or no timestamps, respectively.

As shown in Fig. \ref{fig:ts}, our training data is formatted in two styles, \texttt{<\!|timestamp\!|>} or \texttt{<\!|notimestamp\!|>}. In the \texttt{<\!|timestamp\!|>} case, we add start and end time tokens for each word in the sentence. The timestamps are first obtained by force aligning the speech and ground truth text using NFA \cite{rastorgueva2023nemo}, and then we take the start and end time (in sec) and convert them to frame indices using 80 ms frame rate (determined by the NFA encoder frame rate). Similar to \cite{radford2023robust}, we represent the timestamp tokens as \texttt{<\!|t\!|>}, where \texttt{t} is an integer ranging from 0 to 450, representing a maximum duration of 36 sec. For example, our transcripts with timestamps look like:
\texttt{<\!|\!3\!|\!> classifying <\!|\!14\!|\!> <\!|\!15\!|\!> was <\!|\!16\!|\!> <\!|\!18\!|\!> everything <\!|\!19\!|\!> <\!|\!23\!|\!> to <\!|\!24\!|\!> <\!|\!25\!|\!> him <\!|\!26\!|\!>}. Each timestamp token is tokenized to exactly one ID in the tokenizer output space. To tokenize transcripts, similar to \cite{puvvada2024less}, we use SentencePiece \cite{kudo2018sentencepiece} and concatenated tokenizer \cite{dhawan2023unified} with a vocabulary size of 1024 for each supported language and a sub-tokenizer for the special tokens. Using special tokens enables a one-to-one mapping between the timestamp and the output ID, which helps in reducing the target sequence length. We have also tried to tokenize the timestamps directly using the SentencePiece tokenizer but did not get reasonable performance. Our model is multilingual and we use the same special tokens for timestamps across different languages. We also tried predicting the time shift from the previous timestamp to current timestamp (similar to duration) but did not manage to get good performance. 

In AST training, our AST text translations are generated synthetically using machine translation models from ASR transcripts \cite{puvvada2024less}. Therefore, we have both ASR transcripts and AST translations. To generate timestamps for AST, we first align speech with ASR transcripts using NFA, and then use awesome-align \cite{dou2021word} to align ASR transcripts (source) with AST translations (target) at the word level. Timestamps of each source word are then copied to the corresponding target word. Since awesome-align \cite{dou2021word} shows reasonable zero-shot performance, we use it for aligning between English and three languages: French, German, and Spanish, even though Spanish has not been seen in their training data. We have obtained reasonable performance as shown in the results.

\section{Experiments}
\label{sec:exp}

\subsection{Data and Evaluation Metrics}
\label{sec:data}

We use a subset of \replaced{the Multitask}{Canary} training data \cite{puvvada2024less} to generate timestamps, i.e., 4,275, 1,410, 1,397 and 1,795 hours of speech with word-level timestamps for English, German, Spanish and French, respectively. We use the NeMo Forced Aligner (NFA) \cite{rastorgueva2023nemo} to align the training data for each language separately. To prevent regression, we also use non-timestamped data \cite{puvvada2024less} in training. Overall, the timestamp data accounts for 15\% of all ASR training data. We use equal weights between ASR and AST timestamp training data. We evaluate model performance for ASR timestamp prediction using LibriSpeech test-other \cite{panayotov2015librispeech} and MCV-16.1 \cite{ardila2019common}. For AST, we use FLEURS \cite{conneau2023fleurs} to evaluate our timestamp prediction performance for 6 language pairs, i.e. En$\to$X and X$\to$En, where X is German, French or Spanish. ASR performance is measured using WER, and we use BLEU \cite{post2018call} and COMET \cite{rei2020comet} for AST evaluation.

\begin{table*}[t]
\centering
\begin{tabular}{|c | c c | c | c | c| c |}
\hline
\multirow{2}{*}{Metric} & \multicolumn{2}{c}{En} & \multicolumn{1}{c}{Es} & \multicolumn{1}{c}{Fr} & \multicolumn{1}{c|}{De}  & \multirow{2}{*}{Avg.}\\
 & LS test-other & MCV-16.1 & MCV-16.1 & MCV-16.1 & MCV-16.1 & \\ \hline \hline
Precision (\%) & 93.7 & 83.0 & 93.5 & 85.4 & 90.7 & 89.3 \\ \hline
Recall (\%) & 93.6 & 84.5 & 93.9 & 88.1 & 91.1 & 90.2 \\ \hline
SD (ms) & 50 & 69 & 22 & 22 & 112 & 55 \\ \hline
ED (ms) & 54 & 70 & 42 & 36 & 127 & 66 \\ \hline
\end{tabular}
\caption{Timestamp prediction results in precision \& recall, and start \& end time differences w.r.t. the NFA ground truth.}
\label{table:timestamp}
\vspace{-2em}
\end{table*}

\begin{table*}[h]
\centering
\begin{tabular}{|c| c c | c | c | c | c|}
\hline
\multirow{2}{*}{Model} & \multicolumn{2}{c}{En} & \multicolumn{1}{c}{Es} & \multicolumn{1}{c}{Fr} & \multicolumn{1}{c}{De} & \multirow{2}{*}{Avg.} \\
& LS test-other & MCV-16.1 & MCV-16.1 & MCV-16.1 & MCV-16.1 & \\ \hline \hline
Canary Baseline (\texttt{B0}) & \textbf{4.8} & \textbf{11.7} & \textbf{5.7} & 15.0 & \textbf{7.0} & \textbf{8.8} \\ \hline
\hspace{5mm}+ start \& end TS data (\texttt{E1}) & 5.0 & 12.1 & 5.9 & \textbf{14.8} & 7.1 & 9.0 \\ \hline
\end{tabular}
\caption{WER comparison between the baseline Canary model and the proposed model.}
\label{table:wer}
\vspace{-2em}
\end{table*}

\begin{table}[t]
\centering
\begin{tabular}{|c | c c|}
\hline
\multirow{2}{*}{Error Threshold} & \multicolumn{2}{c|}{Precision / Recall (\%)} \\
& \texttt{E1} & WhisperTimestamped \\ \hline \hline
240 (ms) & \textbf{95.8} / \textbf{95.6} & 83.6 / 83.2 \\ \hline
320 (ms) & \textbf{96.1} / \textbf{95.9} & 92.8 / 92.2 \\ \hline
400 (ms) & \textbf{96.3} / \textbf{96.1} & 95.3 / 94.7 \\ \hline
480 (ms) & \textbf{96.4} / \textbf{96.2} & 96.1 / 95.5 \\ \hline
\end{tabular}
\caption{Comparison of the proposed model and WhisperTimestamped \cite{lintoai2023whispertimestamped} based on the LibriSpeech test-other set.}
\label{table:comp_wt}
\vspace{-3em}
\end{table}

For timestamp evaluation, we use two types of metrics: 1) Precision and recall, and 2) start-time difference (SD) and end-time difference (ED). For precision \& recall, as in \cite{rastorgueva2023nemo}, we compute the true positive (TP) as number of correctly predicted timestamped word in the hypothesis, false positive (FP) as the incorrectly predicted timestamped words in the hypothesis, and false negative (FN) as the number of timestamped words in the reference but missing or incorrectly predicted in the hypothesis. The precision and recall are then calculated as $precision=\frac{TP}{(TP+FP)}$ and $recall=\frac{TP}{(TP+FN)}$. We define a ``timestamped word" as a predicted word with estimated start and end timestamps, and we consider a timestamped word to be correct if: 1) Exact match of the words between a hypothesis and reference, and 2) Both start and end timestamps differ by less than 240 ms. The chosen threshold of 240 ms (instead of 200 ms in \cite{rastorgueva2023nemo}) is because our encoder has a encoding rate of 80 ms.

We further use start and end time differences (SD \& ED) to quantify the accuracy of the predicted timestamps. Here, we only calculate the SD and ED differences when there is a word match between the hypothesis and reference. The absolute differences are calculated no matter whether the difference is less or more than the 240 ms threshold.

\subsection{Training Details}

We use a 170M \replaced{Multitask}{Canary L} model \cite{puvvada2024less} in our experiments. The model consists of a 17-layer conformer encoder, with a model dimension of 512 and projection dimension of 2048. The decoder is a regular 17-layer transformer decoder \cite{vaswani2017attention} with 512-dimensional layers. A downsampling factor of 8 is used to speed up decoding. We use fixed positional embedding for the decoder. In training, we initialize from an existing checkpoint of a \replaced{Multitask}{Canary L} model \cite{puvvada2024less} and continue to fine tune the whole model. Due to the increased timestamp tokens, we do not initialize the softmax and the embedding layers from the checkpoint. Our input features are 128-dim log-mel features with 10-ms frame rate. Lhotse \cite{zelasko2021lhotse} is used for dataloading with a batch duration of 600 sec per GPU. For model training, we use AdamW optimizer and inverse square root annealing learning rate scheduler with a learning rate of 1e-3 and a warm-up step of 2.5k. \replaced{T}{We implement the model with PyTorch using the NeMo Tookit \cite{kuchaiev2019nemo}, and t}he model is trained on 32 A100 (80G) GPUs for 30k steps, with a batch duration of 600 sec per GPU. In training, we simply sum over the loss for word and timestamp tokens without any priors in weighting. 

\section{Results}
\label{sec:results}

\subsection{Timestamp Prediction for ASR}
\label{sec:ablations}

\begin{table*}[h!]
\centering
\begin{tabular}{|c | c c c | c c c | c|}
\hline
\multirow{2}{*}{Metric} & \multicolumn{7}{c|}{Timestamp Prediction Errors} \\
 & De$\to$En & Fr$\to$En & Es$\to$En & En$\to$De & En$\to$Fr & En$\to$Es & Avg \\ \hline \hline
SD (ms) & 323 & 94 & 173 & 228 & 223 & 181 & \textbf{204} \\ \hline
ED (ms) & 364 & 115 & 199 & 232 & 236 & 198 & \textbf{224} \\ \hline
\end{tabular}%
\caption{Timestamp prediction performance for speech translation using the FLEURS test set.}
\label{table:timestamp_ast}
\vspace{-1.5em}
\end{table*}

\begin{table*}[h!]
\centering
\begin{tabular}{|c | c c c | c c c | c|}
\hline
\multirow{2}{*}{Model} & \multicolumn{7}{c|}{BLEU / COMET} \\
 & De$\to$En & Fr$\to$En & Es$\to$En & En$\to$De & En$\to$Fr & En$\to$Es & Avg \\ \hline \hline
B0 & 29.0 / 80.9 & 28.3 / 81.1 & 20.1 / 79.6 & 25.2 / 74.2 & 34.5 / 76.6 & 19.6 / 77.1 & \textbf{26.1} / \textbf{78.3} \\ \hline
E1 & 23.4 / 78.4 & 22.9 / 79.2 & 16.3 / 77.7 & 20.3 / 71.3 & 30.6 / 73.3 & 16.5 / 74.3 & 21.7 / 75.7 \\ \hline
\end{tabular}%
\caption{Translation performance regression in BLEU and COMET using the FLEURS test set.}
\label{table:timestamp_ast_regression}
\vspace{-1.5em}
\end{table*}

\begin{table*}[h!]
\centering
\footnotesize
\begin{tabular}{|p{8.5cm}|p{8cm}|}
\hline
Predicted German Hypothesis & English Ground Truth \\ \hline \hline
\texttt{<\!|\!0\!|\!>} Jedoch \texttt{<\!|\!21\!|\!>} \texttt{<\!|\!21\!|\!>} aufgrund25 \texttt{<\!|\!28\!|\!>} der \texttt{<\!|\!29\!|\!>} \texttt{<\!|\!29\!|\!>} langsamen \texttt{<\!|\!33\!|\!>} \texttt{<\!|\!34\!|\!>} Kommunikationskanäle \texttt{<\!|\!43\!|\!>} \texttt{<\!|\!43\!|\!>} könnten \texttt{<\!|\!55\!|\!>} \texttt{<\!|\!55\!|\!>} Stile \texttt{<\!|\!62\!|\!>} \texttt{<\!|\!64\!|\!>} im \texttt{<\!|\!65\!|\!>} \texttt{<\!|\!67\!|\!>} Westen \texttt{<\!|\!71\!|\!>} \texttt{<\!|\!91\!|\!>} um \texttt{<\!|\!92\!|\!>} \texttt{<\!|\!95\!|\!>} fünfundzwanzig \texttt{<\!|\!96\!|\!>} \texttt{<\!|\!104\!|\!>} bis \texttt{<\!|\!105\!|\!>} \texttt{<\!|\!106\!|\!>} dreißig \texttt{<\!|\!110\!|\!>} \texttt{<\!|\!111\!|\!>} Jahre \texttt{<\!|\!112\!|\!>} \textbf{\texttt{<\!|\!79\!|\!>} hinterherbleiben \texttt{<\!|\!83\!|\!>}} & \texttt{<\!|\!0\!|\!>} However, \texttt{<\!|\!20\!|\!>} \texttt{<\!|\!23\!|\!>} due \texttt{<\!|\!26\!|\!>} \texttt{<\!|\!26\!|\!>} to \texttt{<\!|\!28\!|\!>} \texttt{<\!|\!28\!|\!>} the \texttt{<\!|\!29\!|\!>} \texttt{<\!|\!30\!|\!>} slow \texttt{<\!|\!33\!|\!>} \texttt{<\!|\!34\!|\!>} communication \texttt{<\!|\!42\!|\!>} \texttt{<\!|\!43\!|\!>} channels, \texttt{<\!|\!52\!|\!>} \texttt{<\!|\!55\!|\!>} styles \texttt{<\!|\!62\!|\!>} \texttt{<\!|\!64\!|\!>} in \texttt{<\!|\!65\!|\!>} \texttt{<\!|\!65\!|\!>} the \texttt{<\!|\!67\!|\!>} \texttt{<\!|\!67\!|\!>} west \texttt{<\!|\!71\!|\!>} \texttt{<\!|\!73\!|\!>} could \texttt{<\!|\!74\!|\!>} \textbf{\texttt{<\!|\!76\!|\!>} lag \texttt{<\!|\!79\!|\!>} \texttt{<\!|\!81\!|\!>} behind \texttt{<\!|\!82\!|\!>}} \texttt{<\!|\!88\!|\!>} by \texttt{<\!|\!89\!|\!>} \texttt{<\!|\!92\!|\!>} 25 \texttt{<\!|\!97\!|\!>} \texttt{<\!|\!100\!|\!>} to \texttt{<\!|\!101\!|\!>} \texttt{<\!|\!103\!|\!>} 30 \texttt{<\!|\!106\!|\!>} \texttt{<\!|\!108\!|\!>} year \texttt{<\!|\!121\!|\!>} \\ \hline
\end{tabular}%
\caption{The proposed model is capable of learning the syntactic reordering in speech translation.}
\label{table:word_swap}
\vspace{-1em}
\end{table*}

\begin{table*}[h!]
\centering
\footnotesize
\begin{tabular}{|p{8.5cm}|p{8cm}|}
\hline
Predicted English Hypothesis & French Ground Truth \\ \hline \hline
 \texttt{<\!|\!0\!|\!>} The \texttt{<\!|\!1\!|\!>} \texttt{<\!|\!22\!|\!>} surface \texttt{<\!|\!28\!|\!>} \texttt{<\!|\!28\!|\!>} of \texttt{<\!|\!29\!|\!>} \texttt{<\!|\!29\!|\!>} the \texttt{<\!|\!30\!|\!>} \texttt{<\!|\!31\!|\!>} moon \texttt{<\!|\!34\!|\!>} \texttt{<\!|\!35\!|\!>} is \texttt{<\!|\!36\!|\!>} \textbf{\texttt{<\!|\!36\!|\!>} constituted \texttt{<\!|\!42\!|\!>}} \texttt{<\!|\!42\!|\!>} of \texttt{<\!|\!43\!|\!>} \texttt{<\!|\!44\!|\!>} stone \texttt{<\!|\!47\!|\!>} \texttt{<\!|\!48\!|\!>} and \texttt{<\!|\!49\!|\!>} \texttt{<\!|\!49\!|\!>} of \texttt{<\!|\!50\!|\!>} \texttt{<\!|\!50\!|\!>} duster \texttt{<\!|\!56\!|\!>} \texttt{<\!|\!58\!|\!>} its \texttt{<\!|\!60\!|\!>} \textbf{\texttt{<\!|\!60\!|\!>} louch \texttt{<\!|\!64\!|\!>}} \texttt{<\!|\!65\!|\!>} 
 exterior \texttt{<\!|\!71\!|\!>} \texttt{<\!|\!72\!|\!>} is \texttt{<\!|\!73\!|\!>} \texttt{<\!|\!74\!|\!>} called \texttt{<\!|\!77\!|\!>} \texttt{<\!|\!78\!|\!>} the \texttt{<\!|\!79\!|\!>} \textbf{\texttt{<\!|\!80\!|\!>} croute \texttt{<\!|\!87\!|\!>}} & \texttt{<\!|\!0\!|\!>} La \texttt{<\!|\!1\!|\!>} \texttt{<\!|\!22\!|\!>} surface \texttt{<\!|\!28\!|\!>} \texttt{<\!|\!28\!|\!>} de \texttt{<\!|\!29\!|\!>} \texttt{<\!|\!29\!|\!>} la \texttt{<\!|\!30\!|\!>} \texttt{<\!|\!31\!|\!>} lune \texttt{<\!|\!34\!|\!>} \texttt{<\!|\!35\!|\!>} est \texttt{<\!|\!36\!|\!>} \texttt{<\!|\!36\!|\!>} constituée \texttt{<\!|\!42\!|\!>} \texttt{<\!|\!42\!|\!>} de \texttt{<\!|\!43\!|\!>} \texttt{<\!|\!43\!|\!>} pierres \texttt{<\!|\!48\!|\!>} \texttt{<\!|\!48\!|\!>} et \texttt{<\!|\!49\!|\!>} \texttt{<\!|\!49\!|\!>} de \texttt{<\!|\!50\!|\!>} \texttt{<\!|\!50\!|\!>} poussière. \texttt{<\!|\!58\!|\!>} \texttt{<\!|\!58\!|\!>} Sa \texttt{<\!|\!59\!|\!>} \texttt{<\!|\!60\!|\!>} couche \texttt{<\!|\!64\!|\!>} \texttt{<\!|\!65\!|\!>} extérieure \texttt{<\!|\!71\!|\!>} \texttt{<\!|\!73\!|\!>} est \texttt{<\!|\!74\!|\!>} \texttt{<\!|\!74\!|\!>} appelée \texttt{<\!|\!78\!|\!>} \texttt{<\!|\!78\!|\!>} la \texttt{<\!|\!79\!|\!>} \texttt{<\!|\!80\!|\!>} croûte. \texttt{<\!|\!88\!|\!>} \\ \hline
\end{tabular}
\caption{The proposed model shows capability of predicting timestamps for speech translation when trained only on ASR timestamp data.}
\label{table:generalization}
\vspace{-3em}
\end{table*}

\subsubsection{Timestamp Prediction Quality}

As shown in Table \ref{table:timestamp}, we evaluate the timestamp prediction performance of the proposed method. We use precision and recall to measure the overall timestamp prediction accuracy. The NFA alignments generated by ground truth ASR transcripts are used as the references. As we can see from Table \ref{table:timestamp}, our model achieved a precision of 93.7\% for LibriSpeech test-other and 89.3\% on average. The recalls are 93.6\% and 90.2\%, respectively. This shows that our model can accurately capture most words with their timestamps. We then calculate the predicted SD and ED for each word when there is a word match. As shown in Table \ref{table:timestamp}, the SD and ED range from 22 to 127 ms. In particular, the predictions are fairly accurate for English, Spanish, and French, and relatively worse for German. 

\subsubsection{WER Regression}

In addition to timestamp prediction performance, we also want to make sure the model does not degrade performance on ASR transcripts. Therefore, in Table \ref{table:wer}, we compare the WER of the baseline Canary model (\texttt{B0}) and the proposed model with timestamp data in training (\texttt{E1}). The baseline Canary model is trained without any timestamp data, and fine tuned for the same number of training steps as the timestamp model for fair comparison. To calculate the WER For \texttt{E1}, we use the \texttt{<|timestamp|>} to prompt the model but ignore the timestamp prediction errors and only compute the WER of the words. Table \ref{table:wer} shows that there is only small degradation for ASR performance (around 0.2\% on average) compared to baseline. Additionally, we have also tried prompting the model with \texttt{<\!|notimestamp\!|>} to only generate ASR transcripts without timestamps and there is no WER regression at all.

\subsubsection{Comparison}

We have compared to WhisperTimestamped \cite{lintoai2023whispertimestamped} in Table \ref{table:comp_wt} using the LibriSpeech test-other set. For WhisperTimestamped, we use the ``small" model, which has a similar size (244M) to our model (175M). We note that WhisperTimestamped \cite{lintoai2023whispertimestamped} derives timestamps in an unsupervised way by aligning each word to a contiguous number of speech frames using dynamic time warping (DTW) based on cross-attention weights, and thus have a single timestamp between two words. Therefore, we use that timestamp as the start timestamp of current word as well as the end timestamp of the previous word. Due to the formatting, the derived start and end timestamps from WhisperTimestamped may inherently differ from the NFA produced ``ground truth", and this should not be penalized in evaluation. We therefore do not calculate the absolute differences between the predicted and reference start and end timestamps but present precision and recall percentages. To minimize differences in ground truth labeling, we also vary the match threshold from 240 to 480 ms to calculate precision and recall for both models. When the threshold is 240 ms, we see that our model performs significantly better than WhisperTimestamped. As we increase the threshold (i.e., relaxing the ground truth labeling convention), Table \ref{table:comp_wt} shows that our model (\texttt{E1}) consistently outperforms WhisperTimestamped across all thresholds.

\subsection{Timestamp Prediction for AST}

We use FLEURS \cite{conneau2023fleurs, puvvada2024less} to evaluate the timestamp prediction performance for AST. We initially follow the steps described in Sect. \ref{sec:model} to generate per-word timestamps for evaluation. However, due to the lack of strict word-to-word mapping caused by translation reordering and errors, it is challenging to directly compare word-level timestamps between ground truth and hypotheses. In addition, the timestamp estimation should not be penalized by incorrect text translations. Therefore, instead of using the ground truth translation for the awesome-align we use the hypotheses (removing timestamps) as the target for alignment. The aligned results are used as references in evaluation. In this way, we have one-to-one correspondence for each word, and the start and end timestamps between references and hypotheses.

As shown in Table \ref{table:timestamp_ast}, the task of predicting timestamps for AST is in general more difficult than ASR. The average start and end time errors are 204 and 224 ms, respectively, across 6 En$\to$X and X$\to$En pairs. Among different language pairs, the performance is significantly worse between German and English. This is probably because the different grammatical structure of German (such as syntactic reordering), which makes timestamp prediction more challenging.

In addition, we evaluate the AST quality regression by BLEU and COMET. We use a \texttt{<|timestamp|>} prompt to generate the text with timestamps for \texttt{E1} and then remove timestamps to only evaluate the text translation. We use SacreBLEU \cite{post2018call} for BLEU calculation and the \texttt{Unbabel/wmt22-comet-da} model from HuggingFace \cite{unbabel_wmt22_comet_da} for COMET \cite{rei2022comet} score calculation. We normalize the COMET score to $[0, 100]$ by multiplying the raw score by 100. As shown in Table \ref{table:timestamp_ast_regression}, the translation ability of the timestamp model degrades by 4.4 BLEU points, while the degradation with respect to the COMET score is 2.6 points.

\subsection{Example and Discussion}
In this section, we show some interesting examples of the predicted sentences with timestamps. Table \ref{table:word_swap} shows an example of English speech to German text translation. Note that our model predicts the German verb with timestamps (i.e., \textbf{\texttt{<\!|79\!|>} hinterherbleiben \texttt{<\!|83\!|>}}) at the end of the sentence, with corresponding start and end timestamps in the English speech (i.e., \textbf{\texttt{<\!|\!76\!|\!>} lag \texttt{<\!|\!79\!|\!>} \texttt{<\!|\!81\!|\!>} behind \texttt{<\!|\!82\!|\!>}}) . This reordering of \textbf{hinterherhinken} is due to the subject-object-verb (SOV) structure of German, whereas the model still preserves the timestamps of the original English speech. We note that the timestamps predicted here are no longer monotonically increasing but follow the correspondence of the original English speech. We also note that here two words in English is mapped to one word in German. It would be interesting and worth discussing how the model should handle non-consecutive mapping between words.

Another interesting observation from our training is that a timestamp model trained using only ASR timestamp data demonstrates the capability of predicting timestamps for AST. We present a French-to-English AST example in Table \ref{table:generalization}. 
The predicted timestamps appear quite reasonable when compared with the French ground truth in the second column of Table \ref{table:generalization}. Again, we note that the model has not been trained on any AST training data with timestamps.
Although the timestamps seem reasonable, this model suffers from substantial translation quality regression from 26.1 to 9.7 BLEU points on average. Some common translation errors (bold in the first column) appear to be incorrectly predicting English words to be their French counterparts with similar pronunciation.
Future research may focus on few-shot learning for AST timestamp prediction.

\section{Conclusion}
We propose a data-driven approach for word-level timestamp prediction for a \replaced{Multitask model}{Canary ASR and AST}, leveraging NeMo Forced Aligner (NFA) and awesome-align as teacher models. Our method achieves 80-90\% precision and recall with timestamp errors of 20-120 ms for ASR. Extending to AST, we observed 200 ms timestamp errors with a 4.4 BLEU and 2.6 COMET drop. Our model performs better than WhisperTimestamped in ASR timestamp prediction. We will release the code and checkpoints to support open-source development. Future work includes refining AST timestamping to reduce translation degradation.

\bibliographystyle{IEEEtran}
\bibliography{ref}

\end{document}